\documentclass[letterpaper, 10 pt, conference]{ieeeconf}  

\IEEEoverridecommandlockouts                              

\title{\LARGE \bf
ViTac: Feature Sharing between Vision and Tactile Sensing for Cloth Texture Recognition
}

\author{Shan Luo$^{1,2,3,4}$, Wenzhen Yuan$^{1}$, Edward Adelson$^{1}$, Anthony G. Cohn$^{2}$ and Raul Fuentes$^{3}$
\thanks{$^{1}$Computer Science and Artificial Intelligence Laboratory, Massachusetts Institute of Technology, 32 Vassar St, Cambridge, MA 02139, USA.}
\thanks{$^{2}$School of Computing, University of Leeds, Leeds, Leeds LS2 9JT, UK.}%
\thanks{$^{3}$School of Civil Engineering, University of Leeds, Leeds LS2 9JT, UK.}%
\thanks{$^{4}$Department of Computer Science, University of Liverpool, Liverpool L69 3BX, UK.}%
}

\usepackage{ifpdf}
\usepackage{graphicx}
\ifpdf

\graphicspath{{Figs/}{Figs/PDF/}}
\else
\graphicspath{{Figs/}}
\fi

\usepackage{hhline}

\usepackage{calc}
\let\labelindent\relax
\usepackage{enumitem}

\usepackage{hyperref}
\hypersetup{
	colorlinks = true,
	linkcolor = {blue},
	citecolor = {blue}
}
\usepackage{url}

\usepackage{nomencl}
\usepackage{multirow}

\usepackage{subcaption}

\usepackage{array}
\usepackage{booktabs}

\newcolumntype{P}[1]{>{\centering\arraybackslash}p{#1}}

\usepackage[flushleft]{threeparttable}

\usepackage{cite}

\usepackage{times}
\usepackage{epsfig}
\usepackage{graphicx}
\usepackage{amsmath}
\usepackage{amssymb}
\DeclareMathOperator*{\argmax}{\arg\!\max}

\begin{document}

\maketitle
\thispagestyle{empty}
\pagestyle{empty}

\begin{abstract}

Vision and touch are two of the important sensing modalities for humans and they offer complementary information for sensing the environment. Robots could also benefit from such multi-modal
sensing ability. In this paper, addressing for the first time (to the best of our knowledge) texture recognition from tactile images and vision, we propose a new fusion method named Deep Maximum Covariance Analysis (DMCA) to learn a joint latent space for sharing features through vision and tactile sensing. The features of camera images and tactile data acquired from a GelSight sensor are learned by deep neural networks. But the learned features are of a high dimensionality and are redundant due to the differences between the two sensing modalities, which deteriorates the perception performance. To address this, the learned features are paired using maximum covariance analysis. Results of the algorithm on a newly collected dataset of paired visual and tactile data relating to cloth textures show that a good recognition performance of greater than 90\% can be achieved by using the proposed DMCA framework. In addition, we find that the perception performance of either vision or tactile sensing can be improved by employing the shared representation space, compared to learning from unimodal data.

\end{abstract}

\section{Introduction}
Vision and tactile sensing are two of the main sensing modalities to perceive the ambient world for humans. We employ eyes and hands in a coordinated way to fulfill complex tasks such as recognition, exploration and manipulation of objects: vision perceives the appearance, texture and shape of objects at a certain distance whereas touch enables the acquisition of detailed texture, local shape and other haptic properties through physical interactions. In addition, we have  experience of ``touching to see" and ``seeing to feel". Specifically, when we intend to grasp an object, we are likely to glimpse it first with our eyes to ``feel" its key features, i.e., shapes and textures, and estimate haptic sensations. Such visual features become unobservable after the object is grasped since vision is occluded by the hand and becomes ineffective. In this case, touch sensation distributed in the hand can assist us to ``see" corresponding features. By tracking and sharing these clues through vision and tactile sensing, we can ``see" or ``feel" the object better.  

Research conducted in neuroscience and psychophysics has investigated  sharing between vision and tactile sensing \cite{ernst2002humans}. Visual imagery has been discovered to be involved in the tactile discrimination of orientation in normally sighted humans \cite{zangaladze1999involvement}. The human brain also employs shared models of objects across multiple sensory modalities such as vision and tactile sensing so that knowledge can be transferred from one to another \cite{newell2001viewpoint}. This sharing of information is especially useful when one sense cannot be used. For instance, it has been found that humans rely more on touch when the texture has small details that are difficult to see \cite{lederman2009haptic}.

\begin{figure}
	\centering
	\begin{subfigure}[b]{0.23\textwidth}
		\centering
		\includegraphics[height=2.9cm,width=3cm]{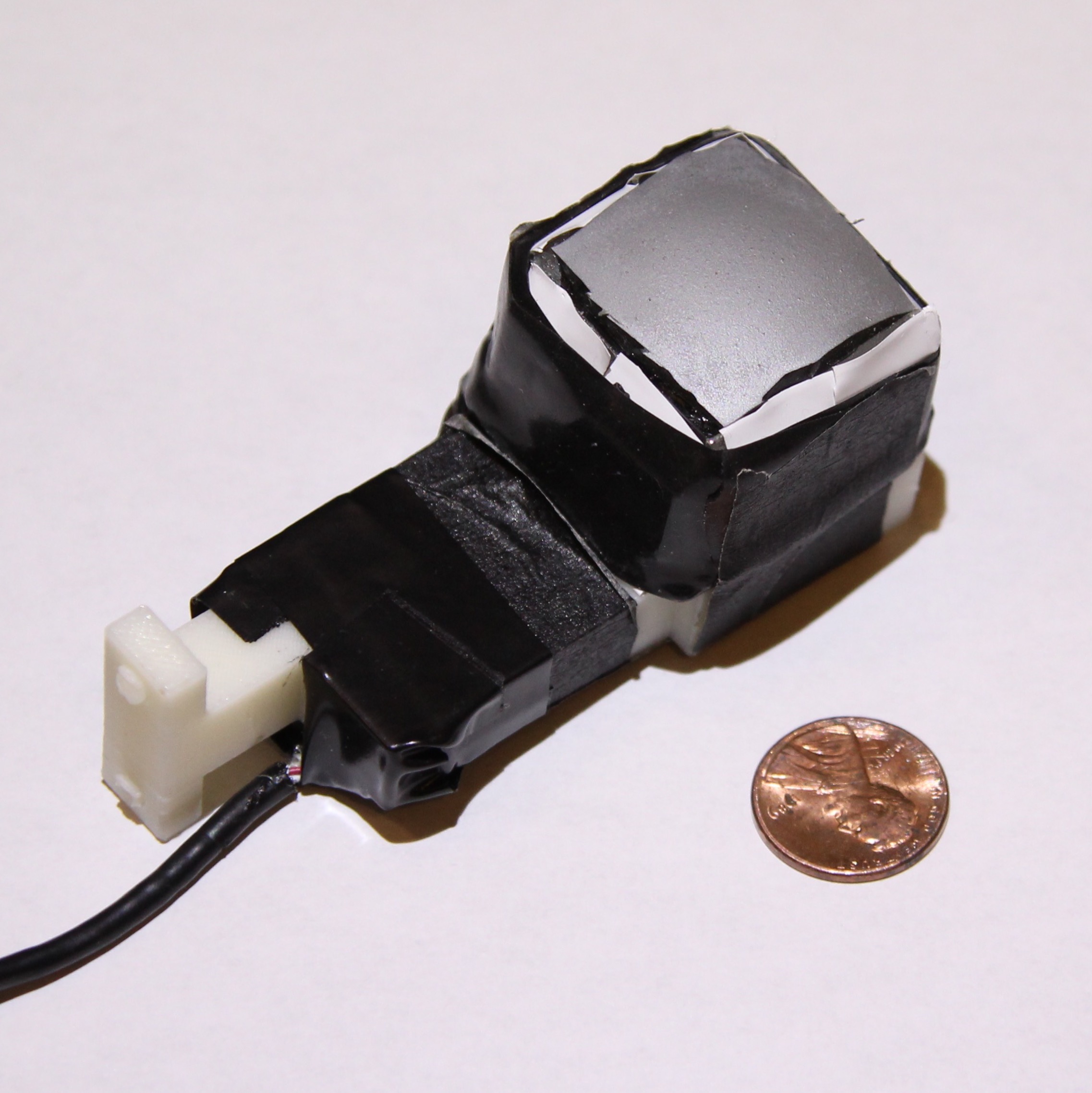}
		\caption{}
		\label{fig:GelSightSensor}
	\end{subfigure}%
	\hfill
	\begin{subfigure}[b]{0.23\textwidth}
		\centering
		\includegraphics[height=2.9cm,width=3.2cm]{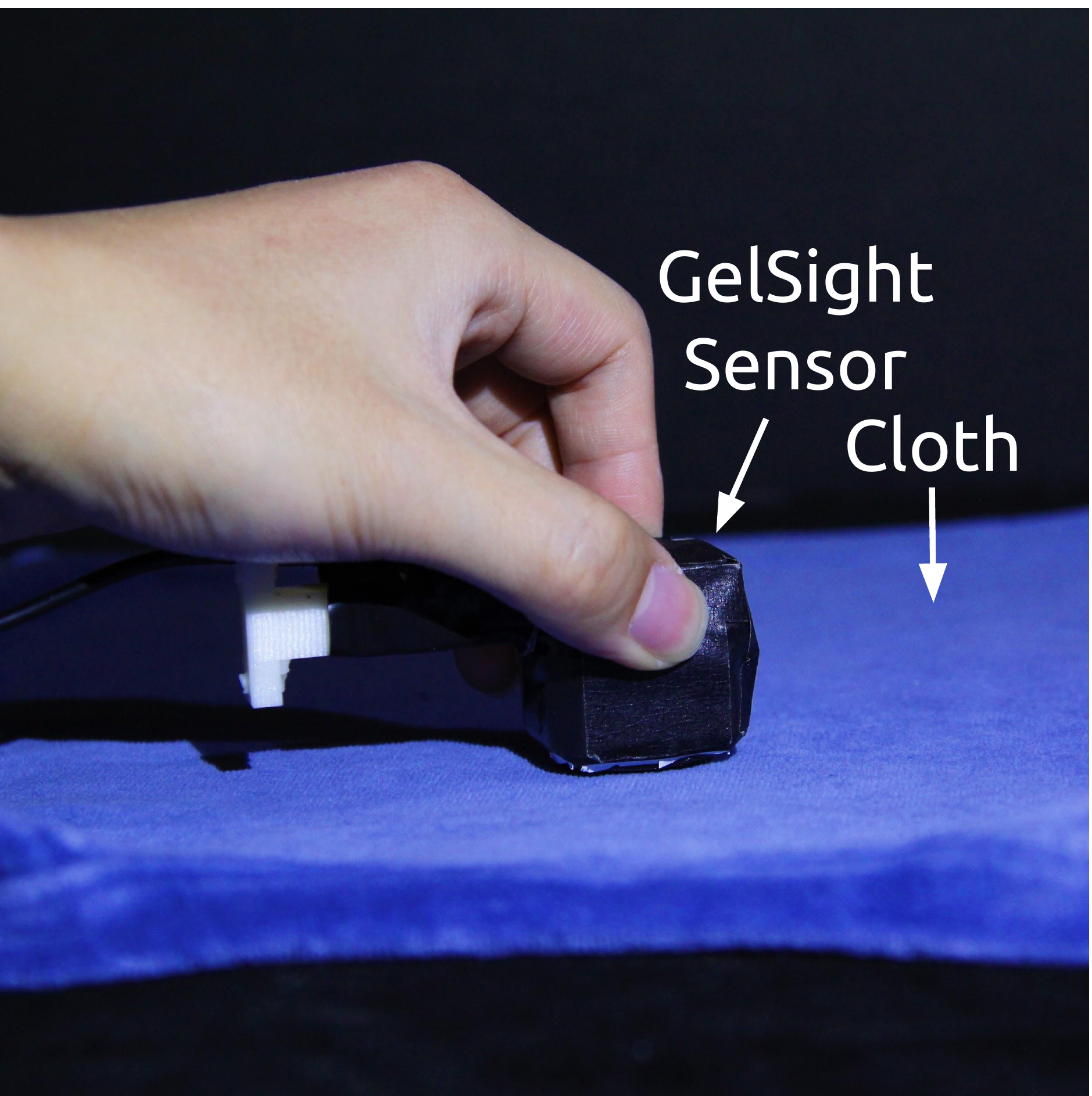}
		\caption{}
		\label{fig:press}
	\end{subfigure}%
	\hfill
	\begin{subfigure}[b]{0.23\textwidth}
		\centering
		\includegraphics[height=2.9cm,width=3.1cm]{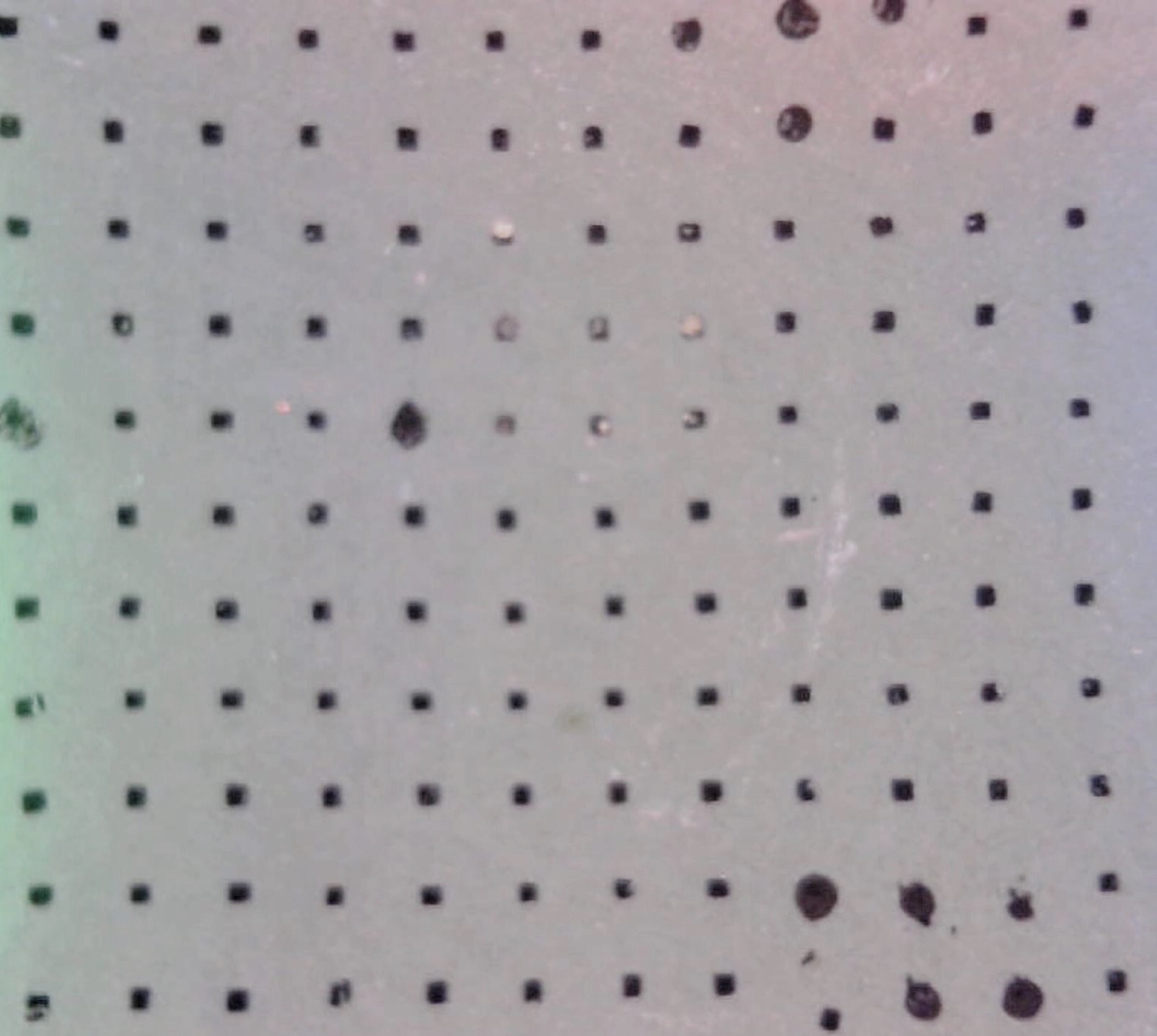}
		\caption{}
		\label{fig:idle}
	\end{subfigure}%
	\hfill
	\begin{subfigure}[b]{0.23\textwidth}
		\centering
		\includegraphics[height=2.9cm,width=3.1cm]{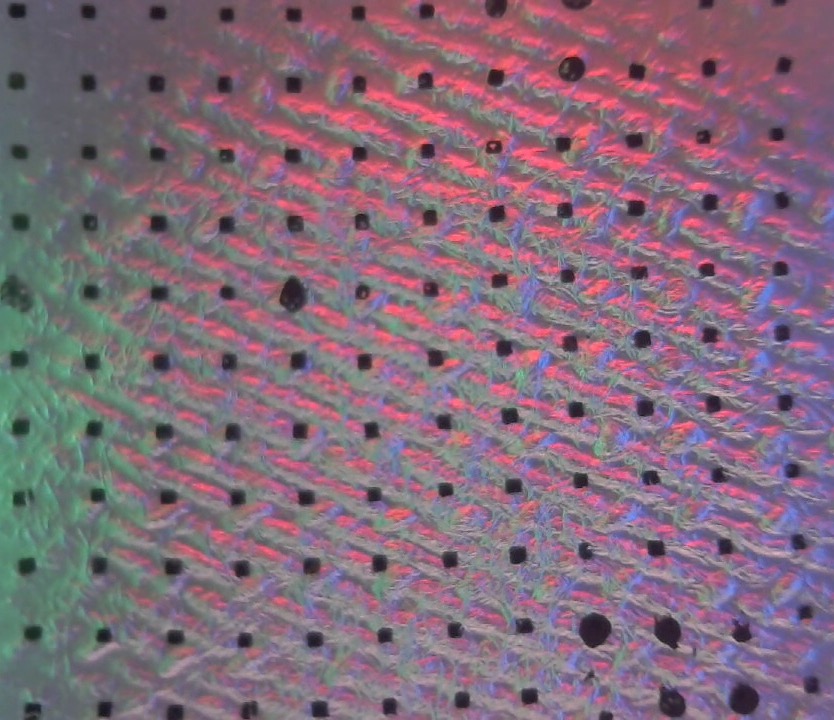}
		\caption{}
		\label{fig:pressGel}
	\end{subfigure}%
	\caption{(a) The GelSight tactile sensor  \cite{johnson2011microgeometry}. (b)  GelSight data is collected by pressing the sensor on clothes. (c) The GelSight image with markers (block dots) captured when no object in contact with the sensor. (d) The GelSight image collected when the coated membrane is deformed by the cloth texture.} 
	\label{fig:gelsight}
\end{figure}

Inspired by the synthesis of vision and tactile sensing in humans, we apply the representation sharing across the two modalities in the artificial perception. There are differences between vision and tactile sensing. For vision, the field of view (FoV) is large and global; there are factors that can affect visual perception, e.g., scaling, rotation, translation, color variance and illumination.In contrast, for tactile sensing, the FoV is small and local as direct sensor-object interactions need to be made; the influence of scaling is shielded as the real dimension and shape of the  object interacted with can be mapped to the tactile sensor directly, whereas the impact of rotation and translation remains. In addition, the different impressions of object shapes caused by forces of various magnitudes and directions resemble the variety of light and illumination conditions in vision. How to learn a joint latent space for sharing features through vision and tactile sensing while eliminating or mitigating the differences between these two modalities is the key issue we will investigate in this paper.

We take cloth texture recognition as the test arena for our algorithms as it is a perfect scenario for sharing features through vision and tactile sensing: the tactile sensing can perceive very detailed texture such as yarn distribution pattern in the cloth whereas vision can capture similar texture pattern (though sometimes is quite blurry). There are also factors that only exist in one modality that may deteriorate the recognition performance. For instance, color variance of cloth is present in vision but is not demonstrated in tactile sensing. We aim to extract the shared information of both modalities while eliminating these factors. 

In this paper, we propose a novel deep fusion framework based on deep neural networks and maximum covariance analysis to learn a joint latent space of vision and tactile sensing. We also introduce a newly collected dataset of paired visual and tactile data. The rest of paper is organized as follows:  related work is reviewed in Section \ref{relatedwork}; the tactile sensor GelSight is introduced in Section~\ref{sensor}; the new dataset combining vision and tactile data is presented in Section~\ref{datacollection}; the proposed framework is illustrated in Section \ref{methodology}; experimental results are presented in Section \ref{experimentresults}; conclusions and future work are described in Section \ref{conclusion}.

\section{Related works} \label{relatedwork}
\subsection{Fusion of vision and tactile sensing}
With attempts dated back to the 1980's \cite{allen1984surface}, tactile sensing has been acting a supporting role for vision in most previous works due to the low resolution of tactile sensors \cite{luo2017robotic,xie2013fiber,luo2015localizing}. By using the tactile device to confirm the object-sensor contact, in \cite{bjorkman2013enhancing} visual features are extracted first to form an initial hypothesis of object shapes and tactile measurements are then used to refine the object model. Hand-designed features can also be extracted from tactile data to form a feature set with visual features. In \cite{bekiroglu2013probabilistic}, image moments of tactile data (both 2D and 3D) are utilized together with vision and action features to create a feature set to facilitate grasping tasks. In \cite{guler2014s}, to identify the content in a container by grasping, the general container deformation is observed by vision and the pressure distributions around the contact regions are captured by tactile sensors. The knowledge embedded in vision and tactile sensing can also be transferred from one to the other, for instance, in \cite{kroemer2011learning} vision and tactile samples are paired to classify materials. In more recent works \cite{gao2015deep} and \cite{pinto2016curious}, deep neural networks are used to extract adjectives/features from both vision and haptic data. Differently from the prior works using low-resolution tactile sensors (for instance a Weiss tactile sensor of 14$ \times $6 taxels used in \cite{luonovel,luo2016iterative}), we use a high-resolution GelSight sensor of (320$ \times $240) to capture more detailed textures. The GelSight sensor is also used in \cite{izatt2017tracking} to fuse vision and touch data where the goal is rather to reconstruct a point cloud representation  and there is no learning of the key features of the two modalities.

\subsection{Multi-modal deep learning}
This work is broadly inspired by the emerging efforts put into learning latent features from multiple modalities. Many works have attempted to investigate the cross-modal relations between vision and sound (especially speeches) modalities \cite{ngiam2011multimodal}. One interesting example is \cite{owens2016visually} where sounds are produced from image sequences. The correlations between other modalities can also be learned from the features using deep neural networks, for example, matching images and caption texts \cite{yan2015deep}, and inferring mechanical properties of fabrics from depth and touch data \cite{yuan2017connecting}. In this work, we leverage the natural synthesis of vision and tactile sensing to learn deep shared representations of these two modalities.

\subsection{Surface texture recognition}
Most previous works on texture recognition employ data from  either vision or tactile sensing only. The most popular hand-crafted features for texture recognition are based on Local Binary Pattern (LBP) descriptors that have been applied in either visual \cite{ojala2002multiresolution} or tactile \cite{li2013sensing} texture recognition. 
In tactile sensing, it is common to move a high-frequency dynamic sensor across an object surface: by sensing the friction arising from the contact, the surface textures can be recognized \cite{johnsson2011sense}. In such works, single-point contact sensors, e.g., whisker-like sensors, are commonly used. In this paper, a high-resolution GelSight sensor is held stationary to press on the texture, which is a much harder problem than the standard texture recognition approaches using a dynamic contact sensor.  Furthermore, to the best of the authors' knowledge, this is the first work to explore both tactile images and vision data for texture recognition.

\section{GelSight touch sensor}\label{sensor}
The GelSight tactile sensor used in this paper is a high-resolution tactile sensor that can capture the surface geometry and texture of interacted objects. It consists of a camera at the bottom and a piece of elastometric gel coated with a reflective membrane on the top, as shown in the Fig.~\ref{fig:GelSightSensor}. The elastomer deforms to take the surface geometry and texture of the objects that it interacts with. The deformation is then recorded by the camera under illumination from LEDs of R, G, B colors that project from various directions through light guiding plates towards the membrane. In this manner, a 3-dimensional height map of the touched surface can then be reconstructed with a photometric stereo algorithm \cite{johnson2011microgeometry}. 

To make the transparent elastomer sensitive to the contact, after trials, the elastomer made of the silicone rubber XP-565 from Silicones, Inc., with the neo-Hookean coefficient $ \mu $ of 0.145MPa, is found most suitable for the task. We use a webcam of 960$ \times $720 and implement the surface topography recovery system on a Matlab platform with the webcam running at over 10 Hz. On the elastomer membrane there are some specially designed markers (square with side length 0.40mm) that can improve the tactile spatial acuity \cite{cramphorn2017addition}.
The sensor is made with inexpensive materials and can give high spatial resolution. In addition, the sensor is not affected by the optical characteristics of the materials being measured like visual cameras, which allows the capture of a wide range of material surfaces. Furthermore, the use of compliant elastomer gel allows the measurement of rich physical properties of  objects interacted with. 

\begin{figure*}[htbp]
	\centering
	\includegraphics[height=4.6cm,width=18cm]{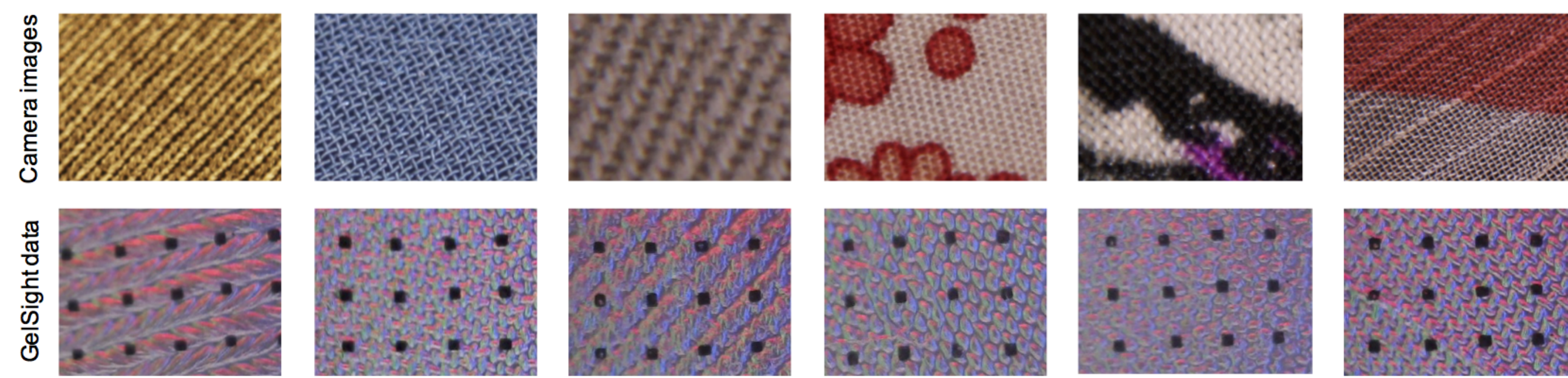}
	\caption{Example camera images (top row) and corresponding GelSight images (bottom row) from the ViTac Cloth dataset. 
To make the textures visually distinguishable, the images shown here are enlarged parts of raw camera/GelSight images.} 
	\label{fig:texturelist}
\end{figure*}

\section{ViTac Cloth dataset}\label{datacollection}
We have built a clothing dataset of 100 pieces of everyday clothing of both visual and tactile data, which we call  the \emph{ViTac Cloth dataset}. The clothing are of various types and are made of a variety of fabrics with different textures. In contrast to available datasets with only either visual images \cite{ojala2002multiresolution} or tactile readings \cite{li2013sensing} of surface textures, the data of two modalities, i.e., vision and touch, was collected while the cloth was lying flat. The color images were first taken by a Canon T2i SLR camera, keeping its image plane approximately parallel to the cloth with different in-plane rotations for a total of ten images per cloth. As a result, there are 1,000 digital camera images in the ViTac dataset. The tactile data was collected by a GelSight sensor. As illustrated in Fig.~\ref{fig:press}, a human holds the GelSight sensor and presses it on the cloth surface in the normal direction. In Fig.~\ref{fig:idle}, a GelSight image with markers is shown as the sensor appears in a non-contact state. As the sensor presses the cloth, a sequence of GelSight images of the cloth texture is captured, as shown in Fig.~\ref{fig:pressGel}. On average each cloth was contacted by the sensor for around 30 times and the number of GelSight readings in each sequence range from 25 to 36. In total 96,536 GelSight images were collected. All the data is based on the shell fabric of the cloth; any hard ornaments on the clothes were precluded from appearing in the view of GelSight or digital camera. Examples of digital camera images and GelSight data
are shown in Fig.~\ref{fig:texturelist}. 

\section{Deep Maximum Covariance Analysis}\label{methodology}
\begin{figure*}[htbp]
	\centering
	\includegraphics[height=3.4cm,width=17cm]{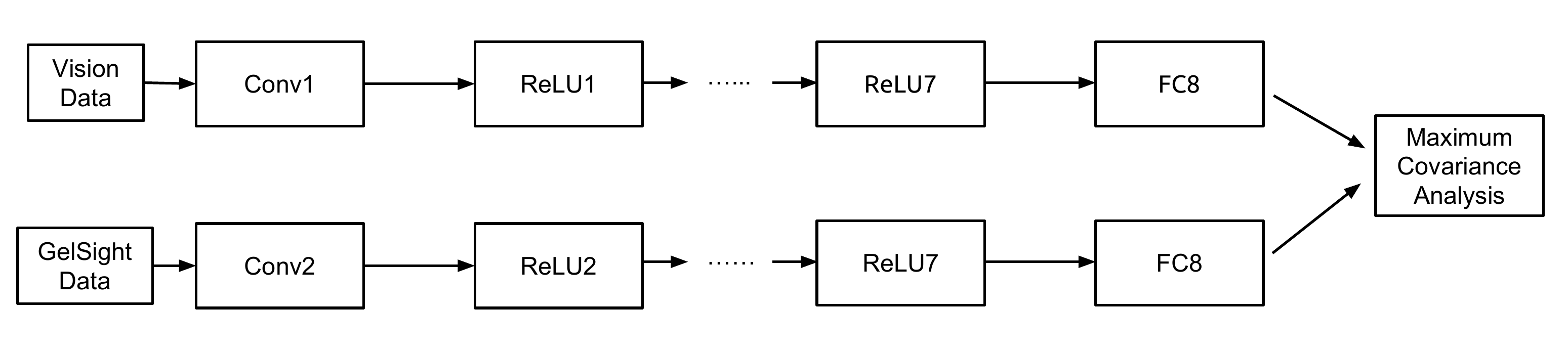}
	\caption{Architecture of the proposed network. The camera and GelSight images are fed into two separate networks and the learned features from the two pipelines are used to achieve a joint latent space using a MCA block.} 
	\label{fig:framework}
\end{figure*}

In this section, we introduce the framework of Deep Maximum Covariance Analysis (DMCA) to match the weakly-paired vision and tactile data. As illustrated in Fig.~\ref{fig:framework}, DMCA first computes representations of the two modalities by passing them through separate multiple stacked layers of a nonlinear transformation and then learns a joint latent space for two modalities such that the covariance between two representations as high as possible.

Let $ X=(x_{1},...,x_{n})\subset\mathbb{R} ^{d \times n} $ and $ X'=(x'_{1},...,x'_{n'})\subset\mathbb{R} ^{d' \times n'} $ be the data sets from two sensing modalities, i.e., camera images and GelSight data, respectively. We aim to obtain the functions $ f $ and $ f' $ to map both $ X $ and $ X' $ to a shared space $ \mathbb{R} ^{q} $. In this multimodal setting, unimodal methods can still be used by just processing each data domain independently such as Principal Component Analysis (PCA). However, the functions $ f $ and $ f' $ can depend on both modalities, therefore, better representations of the information can be retained by finding the dependencies between different modalities than those methods using only unimodal information. There are several methods that can be applied for learning the shared representations of multiple modalities. One typical example is Canonical Correlation Analysis (CCA) that has been used not only for shallowly learned features but also in the context of deep learning \cite{yan2015deep}. But CCA is a distinctive method that constructs lower-dimensional representations suitable for a specific task. It orients the learned features to be projected in a space that discards the information not relevant to this task. In this manner, it can achieve a high performance in one task whereas cannot perform well in another. For this reason, we choose the generative dimensionality reduction method MCA to find lower-dimensional representations of the multimodal data that are appropriate for various tasks. MCA, also known as singular value decomposition (SVD) analysis, constructs a covariance matrix between two datasets and then performs a SVD of the resulting matrix. It is a useful statistical technique for extracting coupled modes of variability between data from two modalities.

Before applying MCA, we learn representations for the two modalities separately to better represent the data from each modality. We feed the camera and GelSight images into two neural networks respectively, as shown in Fig.~\ref{fig:framework}. In this work, the GelSight data is fed into the networks as separate images. In \cite{pinto2016curious}, a ConvNet is built to predict the tactile information during poking given an image of the object and the pretrained CNNs have been used. Following this work, we initialize by pre-training the AlexNet architecture \cite{krizhevsky2012imagenet} and transfer the learned weights for each part of the network. The learned hidden representations from the output of the FC8 layer $ H\subset\mathbb{R}^{D \times n} $ and $ H'\subset\mathbb{R}^{D' \times n'} $ are fed into the MCA layer, where $ D $ and $ D' $ are the dimensions of the hidden representations in the two modalities respectively.

Given two fully paired representation $ H $ and $ H' $, i.e., there is a pairing between each $h_{i}$ and $h'_{i}$. MCA seeks pairs of linear projections  $ W $, $ W' $ that maximise the covariance of the two views:

\begin{equation}
	\begin{split}
		(W^*, W'^{*}) &= \argmax_{W, W'} \textrm{cov} (W^TH, W'^TH') \\
		&=\argmax_{W, W'} \textrm{tr} [W^THH'^TW']
	\end{split}
	\label{mca}
\end{equation}

As mentioned, MCA is a good method for multimodal dimensionality reduction, but it requires fully paired data that is not the case in many applications. For example, in our situation, the visual and GelSight images cannot be fully paired as they are collected in different phases. As tactile data is attained, the camera vision will be obstructed by the GelSight sensor and the state of the cloth will be changed due to the GelSight sensor-cloth interaction.
Therefore the data from the two modalities cannot be fully paired. To solve this kind of weakly paired situation, we employ a variant of MCA proposed in \cite{kroemer2011learning}. Similar to Eq.~\ref{mca}, we perform multimodal dimensionality reduction by solving a SVD problem with projection matrices $ W $ and $ W' $ and also a $ n \times n'$ pairing matrix $ \varPi $ to pair instances from both modalities:

\begin{equation}
	\begin{split}
		(W^*, W'^{*}, \varPi)
		&=\argmax_{W, W', \varPi} \textrm{tr} [W^TH  \varPi H'^TW']
	\end{split}
	\label{wmca}
\end{equation}
Here, $ \varPi \in \{0,1\}^{n \times n'} $, i.e., the elements of $ \varPi $ are either 1 or 0. If $ \varPi_{i,j}=1 $, it implies a pairing between the $i$th vision sample and the $j$th  tactile sample. Each sample is only paired to at most one sample in the other modality, i.e., $ \sum_{i=1}^{n} \varPi_{i,j}\leq1 $ for all ${j=1,...,n'}$ and $ \sum_{i=1}^{n'} \varPi_{i,j}\leq1 $ for all ${i =1,...,n}$. In this manner, the strong pairings between individual samples in a weakly paired group can be inferred.

As (\ref{wmca}) requires both continuous optimization for $ W $ and $ W'$, and combinatoric optimization for $ \varPi $, therefore, there is no single closed-form solution to this optimization. To solve this problem, alternating maximization is applied. First, assumed that $ \varPi $ is known, SVD can be performed as in (\ref{mca}):
\begin{equation}
	\begin{split}
		(W^*, W'^{*})
		&=\argmax_{W, W'} \textrm{tr} [W^TH  \varPi H'^TW']
	\end{split}
\end{equation}
Second, assuming that $ W $ and $ W' $ are known, i.e.,
\begin{equation}
	\begin{split}
		\varPi^*
		&=\argmax_{\varPi} \textrm{tr} [W^TH  \varPi H'^TW']
	\end{split}
\end{equation}
This corresponds to a linear assignment problem and can be solved using the Jonker-Volgenant algorithm that needs expensive computations, especially for the singular value decompositions. In the application of learning shared representations of vision and tactile sensing, the dimension of learned features required to encode the rich information in camera and GelSight images is in the order of $ 10^3 $, for example, we have $ D=4,096 $ for the hidden representations of camera images. To make DWCA practically applicable to our application, we implement both the feature learning and MCA phases on a GPU with the CUBLAS
and CUSOLVER
libraries distributed as part of NVIDIA's CUDA Programming Toolkit\footnote[1]{https://developer.nvidia.com/cuda-toolkit/} to compute linear algebra subroutines.

\section{Experiments and analysis}\label{experimentresults}
We evaluate the proposed DMCA method on cloth texture recognition using tactile and vision data in the ViTac Cloth dataset. We first perform the standard unimodal classification using training and test data of the same single modality. Then we examine the cross-modal classification performance, i.e., training a model based on one sensing modality while apply the model on data of the other modality. This is based on the assumption that visually similar textures are more likely to have similar tactile texture, and vice versa. Lastly, we consider a shared representation learning setting, which is to learn a shared representation of both modalities that is used to recognize textures with single modality in the test phase.

The data in the two modalities in the ViTac Cloth dataset is split into two parts of a 9:1 ratio for training and test data. As stated earlier, the GelSight data and camera images cannot be fully paired, therefore, we use the weak pairing information of which cloth surface the data is recorded from. For each camera image or GelSight reading, we resize the image to $ 256 \times 256 $ first and then extract the center part of image $ 227 \times 227 $ as the input of the neural networks. To measure the performance of the proposed DMCA method, we use the standard multi-class accuracy as our performance metric. We implement our code in Keras
with a Theano backend\footnote[2]{http://deeplearning.net/software/theano/}.

\subsection{Unimodal cloth texture recognition}
We first perform the classic unimodal recognition task using data of each single modality. Following \cite{pinto2016curious}, we fine-tune the AlexNet model and replace the last layer with a fully connected layer of 100 outputs, where 100 is the number of texture classes. 
We use cross-validation to deal with over-fitting, with a learning rate of 0.001, batch size of 128 and 20 epochs used, and the rest of the experiments follow the same configuration. When we use the data from the GelSight sensor for both training and test set, an accuracy of 83.4\% can be achieved for the cloth texture recognition. And when we take the data from the digital camera for both training and test set, an accuracy of 85.9\% can be obtained. This shows that the feature representations learned by deep networks enable  texture recognition with either modality alone. However, especially for robotics,  training data of a particlar modality is not always easy to obtain. For instance, due to limited options of off-the-shelf high-resolution tactile sensors and the high cost of sensor development,  tactile data for objects is neither commonly available nor easy to collect; also, detailed textures of objects are not always easy to access by digital cameras either. To this end, next we explore the cross-modal cloth texture recognition to train a model using one sensing modality while applying the model on data from the other modality.

\begin{table}
	\centering
		\caption{Texture recognition using unimodal modalities and cross modalities of vision and tactile sensing}
		\label{tab:visionvstactile}
		\begin{tabular}{c | c | c }
			\hline
			Training data & Test data & Recognition accuracy \\
			\hhline{=|=|=}
			Vision & Vision & 85.9\% \\
			\hline 
            Tactile & Tactile & 83.4\% \\
			\hline 
			Vision & Tactile & 16.7\% \\
			\hline 
            Tactile & Vision & 14.8\% \\
			\hline
		\end{tabular}
\end{table}

\subsection{Cross-modal cloth texture recognition}
It is possible to recognize cloth textures using data of one modality 
with the model trained on the other because 
both GelSight and 
camera data are presented 
as image arrays 
and cloth textures appear to be of similar patterns 
in both as shown in Fig.~\ref{fig:texturelist}, which is similar to the case when humans see/feel 
cloth textures. The recognition results of unimodal and cross-modal cloth texture recognition are listed and compared in Table~\ref{tab:visionvstactile}. Perhaps surprisingly, the cross-modal cloth texture recognition performs much worse than the unimodal 
cases. When we evaluate the test data from GelSight sensor using the model trained on vision data, an accuracy of only 16.7\% is achieved. It is even worse when we evaluate the test data from the digital camera using the model trained on GelSight data, only an accuracy of 14.8\% is obtained. The probable reasons 
are factors that make the same cloth pattern appear different in the two modalities. In camera vision, scaling, rotation, translation, color variance and illumination are present. For tactile sensing, impressions of cloth patterns change due to different forces applied to the sensor while pressing. These differences mean that the learned features from one modality may not be appropriate for the other. To extract  correlated features between vision and tactile sensing and preserve these features for cloth texture recognition while mitigating the differences between two modalities, we explore the proposed DMCA method to achieve a shared representation of  textures for both modalities.

\subsection{Shared representation learning for cloth 
recognition} 
In the experiments, we assume that both camera and GelSight data are present during the model learning phase, but only GelSight or camera data is used in the later application to new data. The setting can help us to find whether DMCA can acquire low dimensional representations that demonstrate better information embedded in the bimodal data than those learned from unimodal data. 

\begin{figure}
	\centering
	\includegraphics[height=7cm,width=9.5cm]{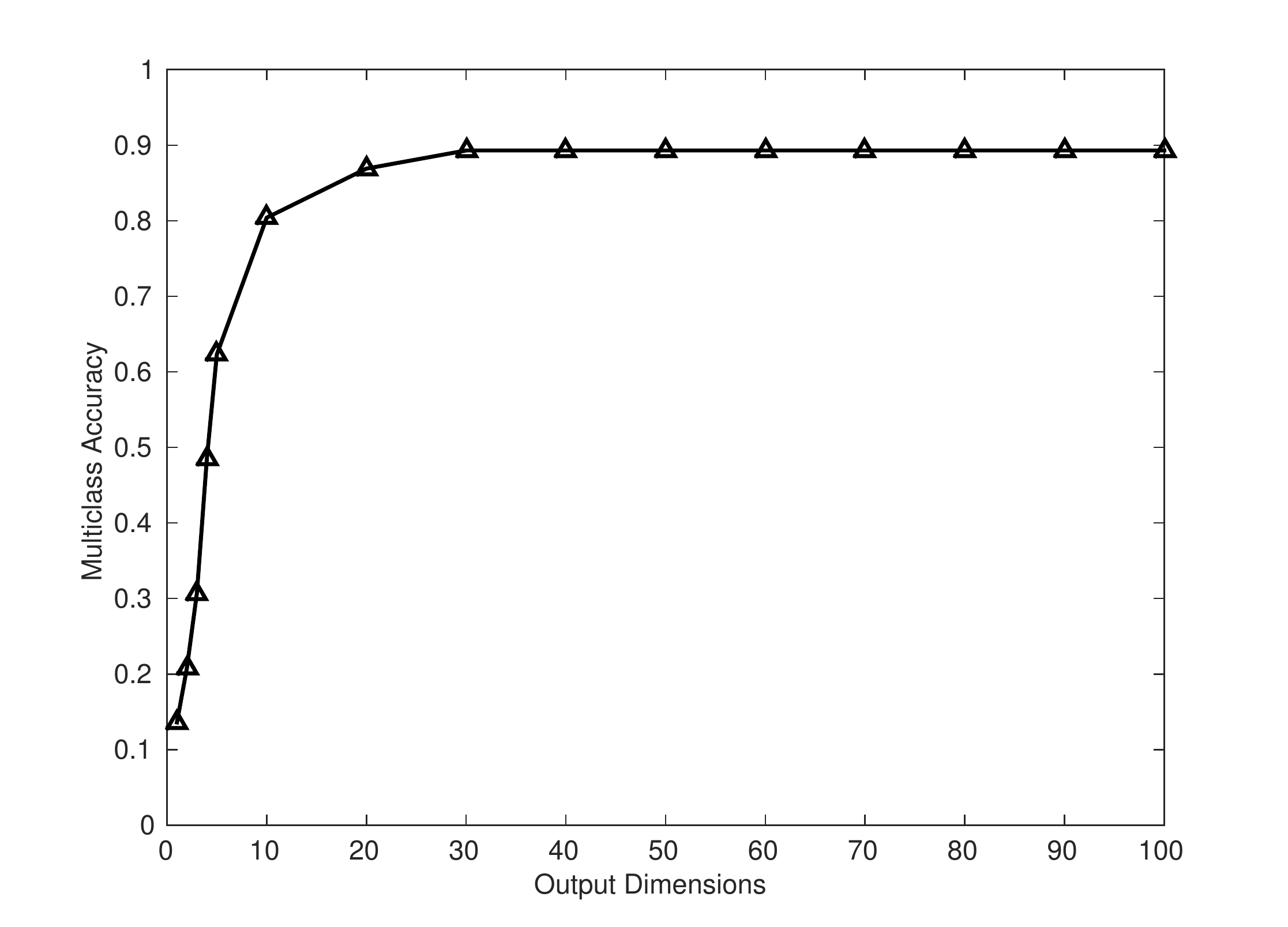}
	\caption{Cloth texture discrimination accuracy with GelSight test data for the different numbers of shared space dimensions, by applying DMCA to bimodal data.}
	\label{fig:tactileclassification}
\end{figure}

We first investigate how the cloth texture classes are classified when only GelSight data is present. As shown in Fig.~\ref{fig:tactileclassification}, the classification performance of DMCA improves as the output dimension becomes larger. As the output dimension continues to increase, the accuracy of DMCA tends to level off and can achieve a classification accuracy of around 90\%. The results show that in DMCA complementary features can be learned from vision to help the tactile modality  discriminate the cloth textures. This is valuable for applying the shared representations in the tactile texture discrimination. As  tactile data of different objects is not easy to  collect due to the high cost of sensor development and time consuming data collection process, it is feasible to add vision data to form a multimodal shared representation with tactile modality so that we can reduce the efforts to collect large volumes of tactile data.


\begin{figure}
	\centering
	\includegraphics[height=7cm,width=9.5cm]{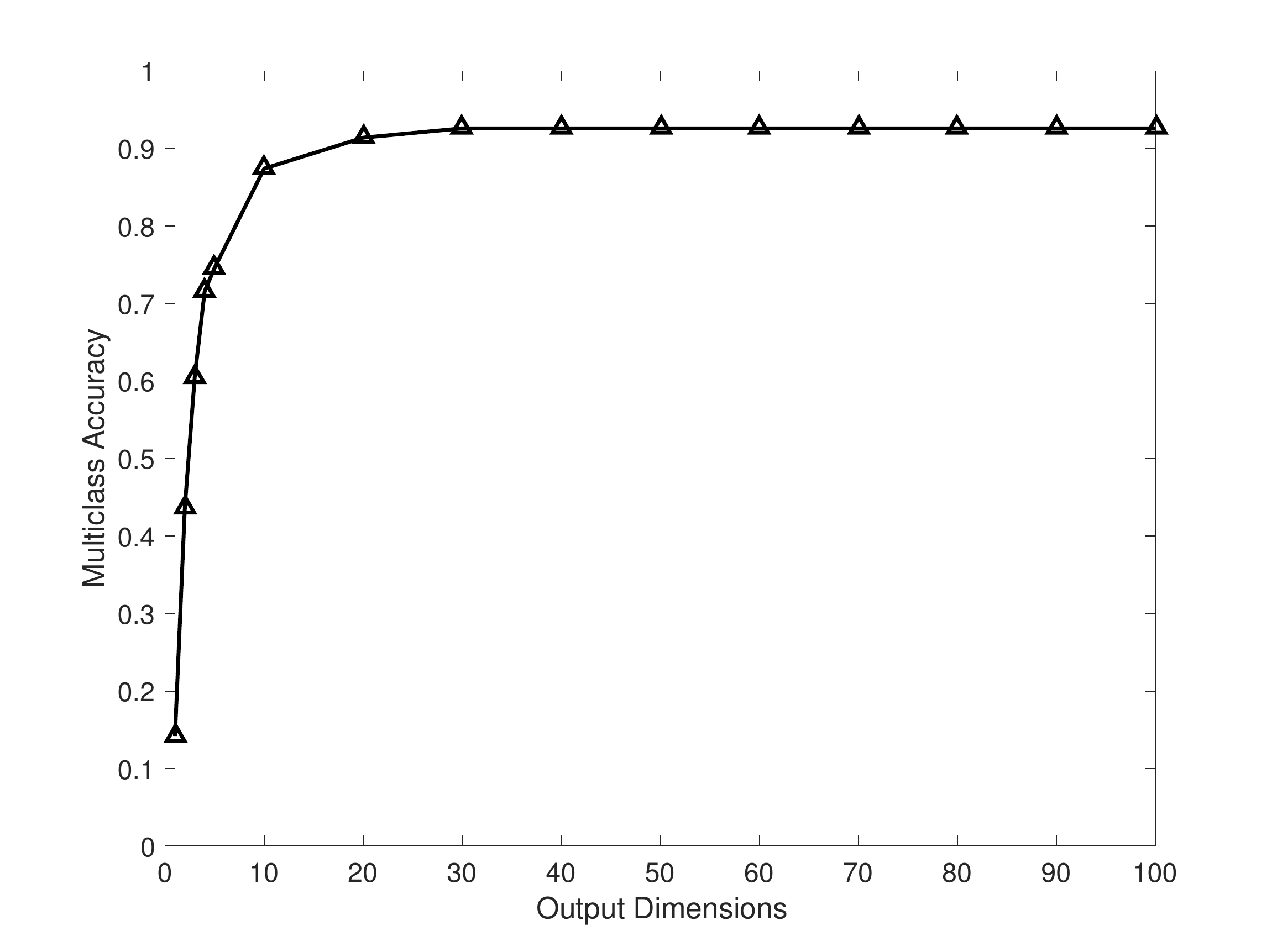}
	\caption{Cloth texture discrimination accuracy with camera test images for the different numbers of shared space dimensions, by applying DMCA to the bimodal data.}
	\label{fig:visualclassification}
\end{figure}

We then look into how the cloth classes can be classified when only camera images are available. As shown in Fig.~\ref{fig:visualclassification}, a similar performance can be observed for DMCA. The  classification performance of DMCA enjoys a dramatic increase as the output dimension increases, and then levels off above dimension20, achievingj a classification accuracy of 92.6\%. The results demonstrate that in DMCA complementary features can also be learned from tactile modality to help vision discriminate cloth textures. 


Overall, the results show that the proposed DMCA learning scheme performs well on the application of tactile-vision shared representations in either tactile or visual cloth texture recognition. This confirms that 
MCA is a powerful tool not only for hand-crafted features \cite{kroemer2011learning}, but also for features learned by deep networks. It has also been demonstrated that inclusion of the other modal data in the learning phase can improve the recognition performance when only one modality is used 
in the test phase.

\section{Conclusion and future work}\label{conclusion}
In this paper, we propose a novel framework for learning joint latent space shared by two modalities, i.e., camera vision and tactile data in our case. To test the proposed framework, a set of experiments was conducted on a newly collected ViTac dataset of both visual and tactile data for a task of cloth texture recognition. Overall, we observe that (1) both vision and tactile sensing modalities can achieve a good recognition accuracy of more than 90\% by using the proposed DMCA method; and (2), the perception performance of either vision or tactile sensing can be improved by employing the shared representation space, compared to learning from unimodal data. There are several directions for  future work:  the  proposed DMCA framework could be applied in other applications, such as learning shared representations from videos, audio soundtracks and subtitles;  or temporal  information  could be included in the latent space. 





\section*{ACKNOWLEDGMENT}

This work was supported by the EPSRC project ``Self-repairing Cities" (EP/N010523/1).


{\small
	\bibliographystyle{ieeetr}
	\bibliography{egbib.bib}
}

\end{document}